\documentclass[letterpaper, 10 pt, conference]{ieeeconf}  

\IEEEoverridecommandlockouts                              

\usepackage{cite}
\usepackage{algorithmic}
\usepackage{graphicx}
\usepackage{mathrsfs}
\usepackage{textcomp}
\usepackage{xcolor}
\usepackage{comment}
\usepackage{amssymb}

\usepackage{amsmath, amsthm}

\usepackage{siunitx}
\usepackage{soul}
\usepackage{multirow}
\usepackage{subcaption}
\usepackage{siunitx}
\usepackage[font=footnotesize]{caption}
\usepackage[colorlinks=true,linkcolor=black,anchorcolor=black,citecolor=black,filecolor=black,menucolor=black,runcolor=black,urlcolor=black]{hyperref}
\usepackage{mathtools}
\usepackage[ruled,vlined,linesnumbered]{algorithm2e}
\usepackage{bm}
\usepackage[normalem]{ulem}
\usepackage{wrapfig}

\usepackage{nomencl}
\usepackage{etoolbox}

\newtheorem{thm}{Theorem}[section]

\newtheorem{prop}[thm]{Proposition}
\newtheorem{prob}[thm]{Problem}

\newtheorem{assum}[thm]{Assumption}

\newtheorem{defn}{Definition}[section]

\newtheorem{rem}{Remark}

\providecommand{\N}{\ensuremath \mathbb{N}}

\newcommand{\set}[1]{\mathcal{#1}}

\newcommand{\state}{\boldsymbol{x}}

\newcommand{\ctrl}{\boldsymbol{u}}

\newcommand{\position}{\boldsymbol{p}}

\newcommand{\ego}{^{\rm ego}}
\newcommand{\ped}{^{p_k}}
\newcommand{\future}{_{[t, t_f]}}
\newcommand{\past}{_{[t_p, t]}}
\newcommand{\loss}{\mathcal{L}}

\newcommand{\nextq}{_{q+1}}
\newcommand{\currq}{_{q}}

\newcommand{\zonotope}{\set{Z}}
\newcommand{\setfn}[1]{\mathscr{#1}}
\newcommand{\zonofnplain}{{\setfn{Z}\!}}
\newcommand{\zonofn}[1]{{\zonofnplain\!\left(#1\right)}}
\newcommand{\centers}{\boldsymbol{c}}
\newcommand{\mink}{^{\rm mink}}

\newcommand{\local}{^{\rm loc}}

\title{\LARGE \bf
Real-time Model Predictive Control with Zonotope-Based Neural Networks for Bipedal Social Navigation}

\author{Abdulaziz Shamsah$^{1,2}$,
Krishanu Agarwal$^{3}$, Shreyas Kousik$^{1,*}$, and Ye Zhao$^{1,*}$
\thanks{$^{1}$George W. Woodruff School of Mechanical Engineering, Georgia Institute of Technology, Atlanta, GA, 30332-0405 USA. \texttt{ashamsah3@gatech.edu}}
\thanks{$^{2}$ Mechanical Engineering Department, College of Engineering and Petroleum, Kuwait University, PO Box 5969, Safat, 13060, Kuwait}
\thanks{$^{3}$School of Electrical and Computer
Engineering, Georgia
Institute of Technology, Atlanta, GA 30308, USA.}
\thanks{$^{*}$Co-senior authorships.}
}

\begin{document}

\maketitle
\pagestyle{empty}


\begin{abstract}
This study addresses the challenge of bipedal navigation in a dynamic human-crowded environment, a research area that remains largely underexplored in the field of legged navigation.
We propose two cascaded zonotope-based neural networks: a Pedestrian Prediction Network (PPN) for pedestrians' future trajectory prediction and an Ego-agent Social Network (ESN) for ego-agent social path planning.Representing future paths as zonotopes allows for efficient reachability-based planning and collision checking. The ESN is then integrated with a Model Predictive Controller (ESN-MPC) for footstep planning for our bipedal robot Digit designed by Agility Robotics. ESN-MPC solves for a collision-free optimal trajectory by optimizing through the gradients of ESN. ESN-MPC optimal trajectory is sent to the low-level controller for full-order simulation of Digit. The overall proposed framework is validated with extensive simulations on randomly generated initial settings with varying human crowd densities.
\end{abstract}

\vspace{0.05in}
\section{Introduction}

Bipedal navigation in complex environments has garnered substantial attention in the robotics community~\cite{huang2023efficient, narkhede2022sequential, zhao2022reactive, Kulgod2020LTL, warnke2020towards}.
Social navigation is a particularly challenging problem due to the inherent uncertainty of the environment, unknown pedestrian dynamics, and implicit social behaviours~\cite{mavrogiannis2023core}. 
Recently, there has been an increasing focus on social navigation for mobile robots in human environments~\cite{moder2022proactive, schaefer2021leveraging, cathcart2023proactive, majd2021safe}.
Nonetheless, the exploration of social navigation for bipedal robots remains largely underexplored.
This can be attributed to the intricate hybrid, nonlinear, and high degrees-of-freedom dynamics associated with bipedal locomotion.

\begin{figure}[t]
\centerline{\includegraphics[width=.8\columnwidth]{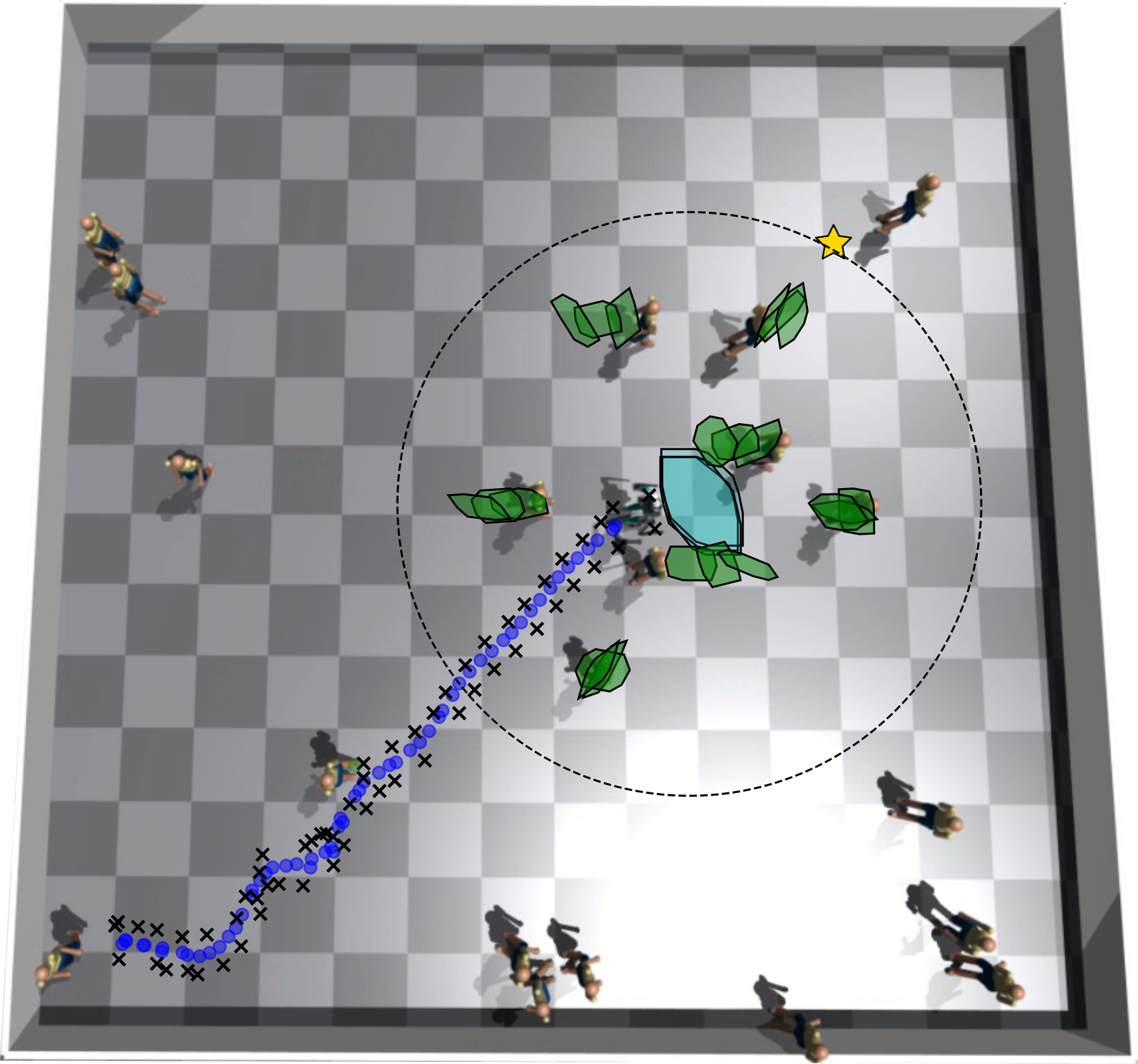}}
\caption{Snapshot of the simulation environment with superimposed zonotopes for the proposed reachability-based social navigation framework. The environment is a $14$ m $\times$ $14$ m open space with $20$ pedestrians.}
\label{fig:high-level}
\end{figure}

In this study, we present an integrated framework for prediction and motion planning for socially acceptable bipedal navigation in human-crowded environments as shown in Fig.~\ref{fig:high-level}.
We propose a navigation framework composed of two cascaded neural networks: a Pedestrian Prediction Network (PPN) for pedestrians' future trajectory prediction and an Ego-agent Social Network (ESN) for ego-agent social path planning.
Our neural networks output reachable sets for pedestrians and the ego-agent represented as zonotopes, a convex symmetric polytope.
Zonotopes offer efficient, yet robust reachability-based planning, collision checking, and uncertainty parameterization~\cite{paparussozapp, selim2022safe, kousik2019safe, althoff2010reachability}.
In this study, we use zonotopes to detect and avoid collisions by checking for intersections between the zonotopes corresponding to the ego-agent and pedestrians.

\begin{figure*}[t]
\centerline{\includegraphics[width=.8\textwidth]{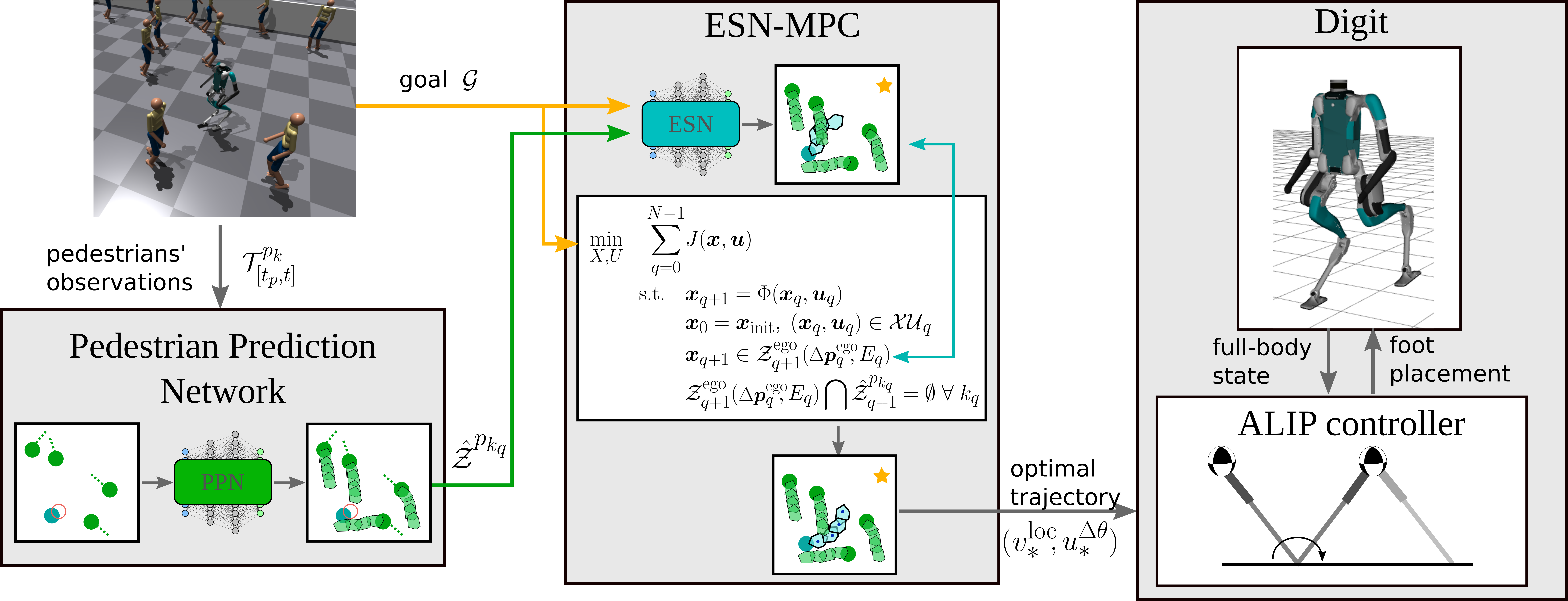}}
\caption{Block diagram of the proposed framework. The framework is composed of two sub-networks: the Pedestrian Prediction Network (PPN) and the Ego-agent Social Network (ESN) shown in green and cyan, respectively (Sec.~\ref{sec:social_zono_net}). 
Given an environment with observed pedestrians and a goal location, PPN predicts the future pedestrians' reachable set.
ESN-MPC optimizes through ESN to generate collision-free trajectories for Digit (Sec.~\ref{sec:SDMPC}).
The optimal trajectory is then sent to the ALIP controller~\cite{Gong2022AngularMomentum} to generate the desired foot placement for reduced-order optimal trajectory tracking.
An ankle-actuated-passivity-based controller~\cite{sadeghian2017passivity,shamsah2023integrated} is implemented on Digit for full-body trajectory tracking.}
\label{fig:block_framework}
\end{figure*}

Our framework integrates ESN in a model predictive controller (MPC) as shown in Fig.~\ref{fig:block_framework}.
The ESN-MPC optimizes over the output of the neural network, with reachability and collision avoidance constraints. ESN-MPC incorporates a reduced-order model (ROM) for the bipedal locomotion process and then sends optimal commands, i.e., center of mass (CoM) velocity and heading change, to the low-level controller on Digit for full-body joint trajectory design and control. 

The main contributions of this study are as follows:
\begin{itemize}
\item A zonotope-based prediction and planning framework for bipedal navigation in a social environment.

\item Novel loss functions to shape zonotopes that represent the future social trajectory of the ego-agent.
\item A framework for hierarchically integrating the neural networks with an MPC and a low-level passivity controller for full-body joint control of Digit.

\end{itemize}

This article is outlined as follows.
Section~\ref{sec:related_work} is a literature review of related work.
Section~\ref{sec:prob_formulation} introduces the problem we are seeking to solve.
Then, the environment setup and zonotope preliminaries are in Section~\ref{sec:prelim}.
Section~\ref{sec:social_zono_net} presents the neural network architecture and loss functions.
Section.~\ref{sec:SDMPC} formulates the problem as an MPC. Implementation details and results are in Section~\ref{sec:implementation}. Finally, Section~\ref{sec:conclusion} concludes this article.
\vspace{0.05in}
\section{Related Work}\label{sec:related_work}

Navigating an environment with humans in a socially compliant manner requires a proactive approach to motion planning \cite{moder2022proactive, schaefer2021leveraging, cathcart2023proactive}.
In~\cite{cathcart2023proactive}, the authors use opinion dynamics to proactively design motion plans for a mobile robot, without the need for human prediction models.
It relies only on the observation of the approaching human position and orientation to form an opinion that alters the neutral path and avoids collisions with pedestrians. Gradient-based trajectory optimization is introduced in \cite{schaefer2021leveraging} to minimize the difference between the humans' future path prediction conditioned on the robot's plan and the nominal prediction. The studies of~\cite{moder2022proactive, schaefer2021leveraging} both assume that a minimally-invasive robot trajectory, with minimal effect on surrounding humans' nominal trajectory, is socially acceptable. In contrast, our work aims to learn the socially acceptable trajectory from human crowd datasets to minimize any heuristic biases on what a socially acceptable trajectory is. 

Our framework is inspired by the human trajectory prediction community~\cite{mangalam2020not, salzmann2020trajectron++, gupta2018social, sadeghian2019sophie}, where we aim to design a socially acceptable trajectory for the ego-agent that mimics the path learned from human crowd datasets.
The work in \cite{hong2023obstacle} proposes an obstacle avoidance learning method that uses a Conditional Variational Autoencoder (CVAE) framework to learn a temporary target distribution to avoid pedestrians actively.
However, during the learning phase, the temporary targets are selected heuristically.
In contrast, we aim to learn such temporary waypoints from human crowd datasets to capture a heuristic-free socially acceptable path.
In~\cite{mangalam2020not}, the authors develop a simple yet, accurate CVAE architecture based on Multi-Layer Perceptrons (MLP) networks to predict crowd trajectories conditioned on past observations and intermediate endpoints. Our ESN follows a similar MLP-based CVAE architecture, where the ego-agent path is conditioned on the final goal location, surrounding pedestrians' future trajectories, and immediate change in the ego-agent state.
Utilizing a non-complex network architecture is pivotal for enabling real-time planning and prediction when integrated into gradient-based motion planning for the ego-agent.

The authors in  \cite{paparussozapp} present a Zonotope Alignment of Prediction and Planning (ZAPP) that relies on zonotopes to enable continuous-time reasoning for planning.
They use trajectron++ \cite{salzmann2020trajectron++} to predict obstacle trajectories as a Gaussian distribution. They construct a zonotope over these distributions, which leads to an overapproximation of the uncertainties. We propose learning these distributions directly as zonotopes, bypassing the initial step of predicting Gaussian distributions for pedestrian motion.
This approach is computationally efficient and facilitates real-time integration with an MPC.

\vspace{0.05in}
\section{Problem formulation}
\label{sec:prob_formulation}
\subsection{Robot Model}
\label{subsec:ROM}
\begin{figure}[t]
\centerline{\includegraphics[width=.8\columnwidth]{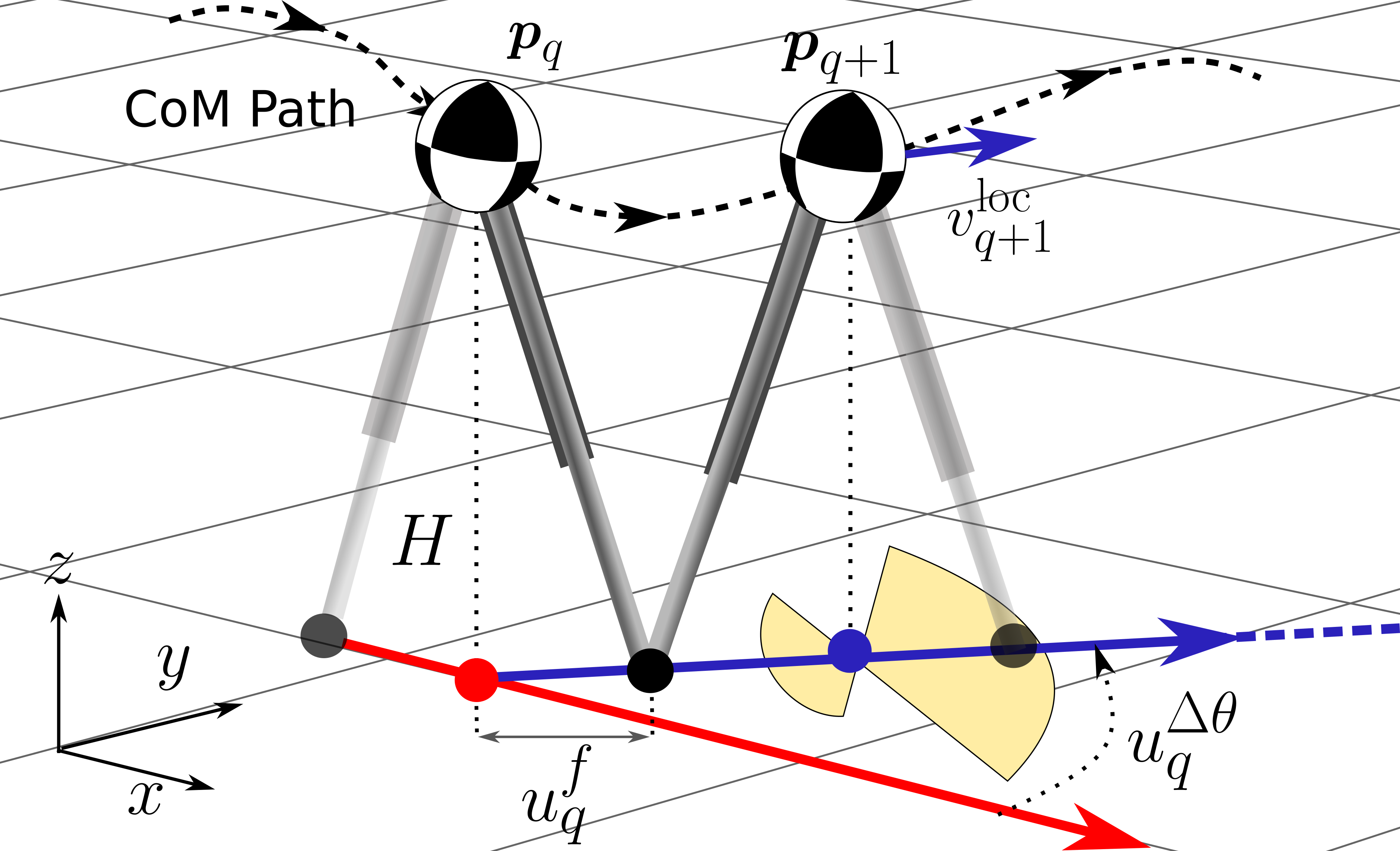}}
\caption{Illustration of the Linear Inverted Pendulum model for two consecutive walking steps, with discrete states $\position \currq$ and $\position \nextq$ at the contact switching time.
The shaded yellow regions indicate the kinematics constraint on the control input $\ctrl$ detailed in Sec.~\ref{subsec:kin_const}.}
\label{fig:LIP}
\end{figure}

Consider a bipedal ego-agent with discrete step-by-step dynamics  $\state\nextq = \Phi(\state\currq, \ctrl \currq)$,
where $\state\currq$ and $\ctrl \currq$ are the state and control input respectively at the contact switching time of the $q^{\rm th}$ walking step.
The robot's state $\state=(\position, v\local, \theta)$, where $\position = (x,y)$ is the 2-D location in the global coordinate, $v\local$ is the local sagittal velocity, and $\theta$ the heading.
The control input is $\ctrl \currq = (u^{f}_{q} \quad u^{\Delta \theta} \currq)$, where $u^{f}_{q}$ is the local sagittal foot position relative to the CoM, and $u^{\Delta \theta} \currq$ is the heading angle change as shown in Fig.~\ref{fig:LIP}.

The reduced-order model (ROM) used to design the walking motion of Digit is the Linear Inverted Pendulum (LIP) model \cite{kajita20013d}. For the LIP model, we assume that each step has a fixed duration $T$\footnote{set to be equal to the timestep between frames in the dataset ($0.4$ s)}~\cite{narkhede2022sequential, teng2021toward}.
Then we build our model on the discrete local sagittal dynamics $(\Delta x\local \currq,v\local\currq)$ \footnote{the lateral dynamics are only considered in the ALIP model at the low level since they are periodic with a constant desired lateral foot placement (See Fig.~\ref{fig:arch})}, where $\Delta x\local=x\local\nextq - x\local\currq$ and $v\local\currq$ is the sagittal velocity at the local coordinate for the $q^{\rm th}$ walking step (see Fig.~\ref{fig:LIP}):

\begin{equation}
    \Delta x\local (u^f\currq) = \left(v\local\currq \frac{\sinh(\omega T)}{\omega} + (1 - \cosh(\omega T))u^f\currq\right)
    \label{eq:delta_x}
\end{equation}
\begin{equation}
    v\local\nextq (u^f\currq) = \cosh(\omega T) v\local\currq - \omega \sinh(\omega T) u^f\currq \cos(\theta_q)
    \label{eq:sagittal_velocity}
\end{equation}
where $\omega = \sqrt{g/H}$, where $g$ is the gravitational constant and $H$ is the CoM height.
Based on the local sagittal dynamics~\eqref{eq:delta_x} and~\eqref{eq:sagittal_velocity}, we add heading angle $\theta \currq $ to control the LIP dynamics in 2-D Euclidean space.
The heading angle change is governed by $\theta\nextq = \theta_{q} + u^{\Delta \theta} \currq$ across walking steps.
Therefore the full LIP dynamics in 2-D Euclidean space become:
\begin{subequations}
\label{eq:lip_dynamics}
\begin{align}
x\nextq &= x_q +\Delta x\local (u^f\currq)\cos(\theta_q) \\
y\nextq &= y_q + \Delta x\local (u^f\currq) \sin(\theta_q) \\
v\local\nextq  &= \cosh(\omega T) v\local\currq - \omega \sinh(\omega T) u^f\currq \cos(\theta_q) \\
\theta\nextq &= \theta\currq + u^{\Delta \theta}\currq 
\end{align}
\end{subequations}

For notation simplicity, hereafter, we  refer to~\eqref{eq:lip_dynamics} as:
\begin{equation}
    \state \nextq = \Phi(\state \currq, \ctrl \currq)
\end{equation}

\subsection{Environment Setup and Problem Statement}
The ego-agent
is tasked to navigate to a known goal location $\set{G}$ in an open environment with $m \in \N$ observed pedestrians treated as dynamic obstacles. 
The pedestrian state $\set{T}\ped\past $ is the 2-D trajectory of pedestrian $k$ observed over the discrete time interval $[t_p, t]$.
The environment is partially observable as only the pedestrians in a pre-specified sensory radius of the ego-agent are observed.
The path the ego-agent takes should ensure 
navigation safety, and promote social acceptability.

\begin{defn}[Navigation safety] Navigation safety is defined as maneuvering in human crowded environments while avoiding collisions with pedestrians, i.e., $\|\position_t -  \set{T}\ped_t \| > d, \; \forall t, k$, where $d$ represent the minimum allowable distance between the ego-agent and the pedestrians.
\label{def:nav_safety}
\end{defn}

\begin{defn}[Socially acceptable path for bipedal systems] A path that a bipedal ego-agent takes in a human-crowded environment is deemed socially acceptable if it has an Average Displacement Error (ADE) $ < \epsilon$~\footnote{$\epsilon$ represents the allowable deviation from the socially acceptable path. The Average Displacement Error denotes the average error between the planned path and the ground-truth path.} when compared to ground truth data in the same environment. 
\label{def:social_path}
\end{defn}

Based on the aforementioned definitions and environment setup the problem we aim to solve is as follows:
\begin{prob}
Given the discrete dynamics of the bipedal robot $\state \nextq = \Phi(\state \currq, \ctrl \currq)$ and an environment state $E=(\set{T}\ped\past,\set{G})$, find a motion plan that promotes social acceptability for the bipedal ego-agent in a partially observable environment containing pedestrians while ensuring 
navigation safety. 
\label{problem_statment}
\end{prob}

\vspace{0.05in}
\section{Preliminaries}
\label{sec:prelim}

To solve the social navigation problem defined above, we propose a learning framework to learn socially acceptable reachable sets parameterized as zonotopes (Sec.~\ref{subsec:learning_arch}).

Problem.\ref{problem_statment} is then reformulated as a step-by-step MPC problem with navigation safety constraints and implemented in real time on our Digit humanoid robot~\cite{agility} (Sec.~\ref{subsec:MPC}).
This section begins by introducing the learning and environment assumptions, and zonotope preliminaries.

\subsubsection{Environment Assumptions and Observations}
In this work, we hypothesize that in a social setting, the information accessible by the ego-agent that is used to determine its future path $\set{T} \ego \future=\{x\ego\currq, y \ego \currq\}^{t_f}_{q=t}$\footnote{the subscripts $t_p$, $t$, and $t_f$ represent a discrete time indices denoting the past, current and future trajectories, respectively, where $t_p < t < t_f$.} are three fold: (i) its final destination $\set{G}=(x^{\rm dest}, y^{\rm dest})$ (ego-agent intent), (ii) the surrounding pedestrians' past trajectory $\set{T} \ped \past=\{x \ped \currq, y \ped \currq\}^{t}_{q=t_p}$ for the $k^{\rm th}$ pedestrian, and (iii) the ego-agent's social experience, i.e., its assumptions on how to navigate the environment in a socially-acceptable manner.
We treat the social experience as latent information that is not readily available in human crowd datasets. Therefore we make the following assumption.

\begin{assum}
Learning the future trajectory of an ego-agent $\set{T} \ego \future$ based on its final goal $\set{G}$ and surrounding pedestrians' past trajectories $\set{T} \ped \past$, will learn the ego-agent's social experience. 
\end{assum}

\begin{figure*}[t]
\centerline{\includegraphics[width=.95\textwidth]{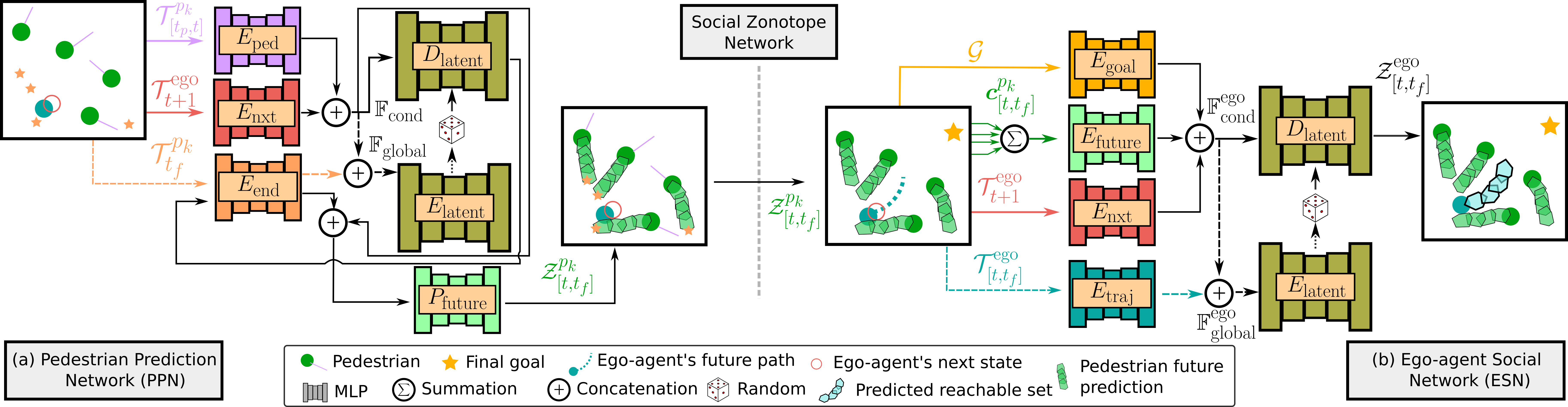}}
\caption{ 
(a) shows the pedestrian prediction network, conditioned on the pedestrian endpoints and the immediate change in the ego-agent's state. (b) shows the ego-agent social network conditioned on the pedestrians' future prediction, the immediate change in the ego-agent's state, and the ego-agent's goal location. Dashed connections are used during training only.}
\label{fig:arch}
\end{figure*}

\subsubsection{Zonotopes Preliminaries}
A zonotope $\zonotope \in \mathbb{R}^n$ is a convex, symmetrical polytope paramterized by a center $\centers \in \mathbb{R}^n$ and a generator matrix $G \in \mathbb{R}^{n \times n_{G}}$ (see Fig.~\ref{fig:zonotope_shaping}).
\begin{equation}
    \zonotope = \zonofn{\centers, G}=\{\centers + G\beta  \; | \; \|\beta\|_\infty \leq 1 \}
\end{equation}

The Minkowski sum of $\zonotope_1 =\zonofn{\centers_1, G_1}$ and $\zonotope_2=\zonofn{\centers_2, G_2}$ is $\zonotope_1 \oplus \zonotope_2 = \mathscr{Z}{(\centers_1+\centers_2, [G_1 \; G_2])}$.
To Check collisions between two zonotopes, \cite[Lemma 5.1]{guibas2003zonotopes} is used:
\begin{prop}(\cite[Lemma 5.1]{guibas2003zonotopes})
\label{prop:intersection}
    $\zonotope_1 \cap \zonotope_1 = \emptyset $ iff $\centers_1 \notin \zonofn{\centers_2, [G_1 \; G_2]}$.
\end{prop}

When $n=2$ zonotopes can be represented as polytopes using the half-space representation $\set{P} = \{x \; | \;Ax \leq b\}$, where $x \in \set{P} \iff \max (Ax - b) \leq 0$ and $ x \notin \set{P} \iff \max (Ax - b) > 0 $.
To convert a 2-D zonotope from the center-generator representation to the half-space representation, we use the following proposition:
\begin{prop}(\cite[Theorem 2.1]{althoff2010reachability})
    Let $C=\begin{bmatrix} - G[2,:]& G[1,:]]\end{bmatrix}$ and $l_{G}[i] = \| G[:,i] \|_2$ the half-space representation of a 2-D zonotope:
    \begin{equation}
        A[i,:] = \frac{1}{l_{G}[i]} \cdot \begin{bmatrix}
C\\
-C
    \end{bmatrix} \in \mathbb{R}^{2 n_G\times 2}
    \end{equation}
\label{prop:half_space}
\begin{equation}
    b = A \cdot c + |{A G}| \; 1_{m\times 1} \in \mathbb{R}^{2 n_G}
\end{equation}
\end{prop}
In this work, zonotopes are used to describe the social reachable set for the ego-agent.
We seek to learn a sequence of social zonotopes $\zonotope \ego \currq$, each of which contains two consecutive waypoints of the ego-agent's future social trajectory $\set{T} \ego \future$.
\begin{defn}[Social Zonotope $\zonotope \ego \currq$]
A social zonotope for the ego-agent's $q^{\rm th}$ walking step is $\zonotope \ego \currq =\mathscr{L}(\centers \currq, G \currq)$, such that $\set{T} \ego \future \in \bigcup\limits_{q=t}^{t_{f}-1} \zonotope \ego_{q}$.  
\end{defn}

\vspace{0.05in}
\section{Social Zonotope Network}
\label{sec:social_zono_net}

\subsection{Learning Architecture}
\label{subsec:learning_arch}

We set up a conditional variational autoencoder (CVAE) architecture to learn the ego-agent's future trajectory conditioned on the final destination goal, the immediate change in the ego-agent's state, and the surrounding pedestrians' past trajectories.
The proposed architecture incorporates Multi-Layer Perceptrons (MLP) with ReLU non-linearity for all the sub-networks.

\subsubsection{Pedestrian Prediction Network (PPN)} The pedestrian prediction network (shown in Fig.~\ref{fig:arch}(a)) is inspired by PECNet~\cite{mangalam2020not}, where the endpoint of the pedestrian trajectory $\set{T} \ped_{t_f}$ is learned first, and then the future trajectory is predicted.
Our proposed network deviates from PECNet in three ways.
First, the pedestrian future trajectory is also conditioned on the immediate change in the ego-agent's state $\set{T} \ego_{t+1}$ (shown in red in Fig.~\ref{fig:arch}(a)).
This coupling of the pedestrian prediction and ego-agent planning networks is intended to capture the effect of the robot's control on the future trajectories of the surrounding pedestrians~\cite{schaefer2021leveraging, mavrogiannis2023core}, and enable bidirectional influence for the entire ego-agent-pedestrian team.
Second, the output of the network is the pedestrian's future reachable set parameterized as zonotopes $\zonotope \ped \future$  rather than trajectories for robust collision checking and uncertainty parameterization~\cite{paparussozapp, selim2022safe, kousik2019safe}.
Third, we replace the social pooling module with a simple ego-agent sensory radius threshold for computational efficiency.

The pedestrians' past trajectories $\set{T} \ped\past$ are encoded in $E_{\rm ped}$ as seen by the purple arrow in Fig.~\ref{fig:arch}(a), while the incremental change in the ego-agent state representing the ego-agent control is encoded in $E_{\rm nxt}$ as seen by the red arrow in Fig.~\ref{fig:arch}(a).
This allows us to condition the prediction of the pedestrians' trajectory on the ego-agent's control.
The resultant latent features $E_{\rm ped}(\set{T} \ped\past)$ and $E_{\rm nxt}(\set{T} \ego_{t +1})$ are then concatenated and used as the condition features $\mathbb{F}_{\rm cond}$.
The pedestrian's endpoint locations are encoded in $E_{\rm end}$ as seen by the orange arrows in Fig.~\ref{fig:arch}(a).
The resultant latent features $E_{\rm end}(\set{T} \ped_{t_f})$  are then concatenated with $\mathbb{F}_{\rm cond}$ as global features $\mathbb{F}_{\rm global}$ and encoded in the latent encoder $E_{\rm latent}$.
We randomly sample features from a normal distribution $\mathcal{N} (\boldsymbol{\mu},\boldsymbol{\sigma})$ generated by the $E_{\rm latent}$ module, and concatenate them with $\mathbb{F}_{\rm cond}$.
This concatenated information is then passed into the latent decoder $D_{\rm latent}$.
Then $D_{\rm latent}$ outputs the predicted endpoint that is passed again through $E_{\rm end}$.
The output is concatenated again with $\mathbb{F}_{\rm cond}$ and passed to $P_{\rm future}$ to output the predicted zonotopes of the pedestrians $\zonotope \ped \future$.

\subsubsection{Ego-agent Social Network (ESN)}

ESN architecture is shown in Fig.~\ref{fig:arch}(b).
The surrounding pedestrians' future zonotope centers $\centers\ped\future$ are aggregated through summation to take into account the collective effect of surrounding pedestrians while keeping a fixed architecture\footnote{Other human trajectory learning modules include a social module to take into account the surrounding pedestrians effect such as social non-local pooling mask~\cite{mangalam2020not}, max-pooling~\cite{gupta2018social}, and sorting~\cite{sadeghian2019sophie}.} \cite{salzmann2020trajectron++}. 
The summed pedestrian features are then encoded in $E_{\rm future}$ as seen by the green arrows in Fig.~\ref{fig:arch}(b). 
The goal location for the ego-agent is encoded in $E_{\rm goal}$, while the incremental change in the ego-agent state is encoded in $E_{\rm nxt}$ as seen by the orange and red arrows respectively in Fig.~\ref{fig:arch}(b).
The resultant latent features $E_{\rm future}(\sum_{k=1}^{m}\centers\ped_{[t,t_f]})$, $E_{\rm goal}(\mathcal{G})$ and $E_{\rm nxt}(\set{T} \ego_{t +1})$ are then concatenated and used as the condition features $\mathbb{F}\ego_{\rm cond}$ for the CVAE.
The ground truth of the ego-agent's future trajectory $\set{T} \ego\future$ is encoded in $E_{\rm traj}$ as shown by the cyan arrows in Fig.~\ref{fig:arch}(b).
The resultant latent features $E_{\rm traj}(\set{T} \ego\future)$ are then concatenated with $\mathbb{F}\ego_{\rm cond}$ as global features $\mathbb{F}\ego_{\rm global}$ and encoded in the latent encoder $E_{\rm latent}$.
Similarly, we randomly sample features from a normal distribution $\mathcal{N} (\boldsymbol{\mu},\boldsymbol{\sigma})$ generated by the $E_{\rm latent}$ module, and concatenate them with $\mathbb{F}\ego_{\rm cond}$.
This concatenated information is then passed into the latent decoder $D_{\rm latent}$, resulting in our prediction of the ego-agent's future reachable set $\zonotope \ego\future$.

\begin{rem}
    Including $E_{\rm nxt}$ in both neural networks facilitates seamless integration with a step-by-step MPC, as the MPC's decision variables ($\Delta\position\ego$) will be used as inputs to $E_{\rm nxt}$ as detailed in Sec.~\ref{sec:SDMPC}.
\end{rem}


\subsection{Zonotope Shaping Loss Functions}
\label{subsec:zonotope_loss}
\begin{figure}[t]
\centerline{\includegraphics[width=.75\columnwidth]{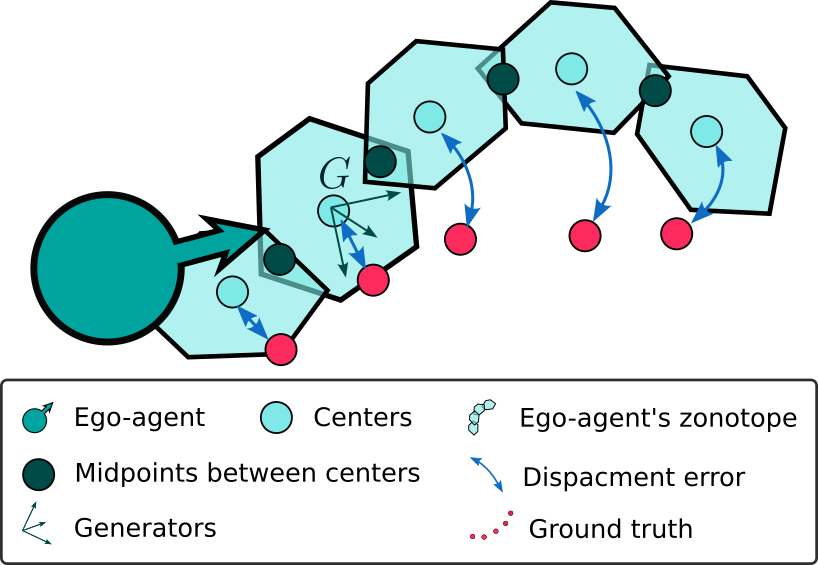}}

\caption{Our zonotope shaping loss functions. The loss aims to learn interconnected zonotopes that engulf the ground truth path.}
\label{fig:zonotope_shaping}
\end{figure}

The zonotope shaping loss functions are used for both PPN and ESN, where both outputs are parameterized as zonotopes.
The goal of these loss functions is three folds: (i) penalize deviation of the centers of the zonotopes from the ground truth future trajectory; (ii) generate intersecting zonotopes for consecutive walking steps; and (iii) reduce the size of the zonotopes to avoid unnecessary, excessively large zonotopes.
Based on these goals, the following is a list of loss functions to shape the zonotopes (Fig.~\ref{fig:zonotope_shaping}):
\begin{enumerate}
    \item Average displacement error between the predicted centers and midpoint of the ground truth trajectory $\mathcal{T}_{{\rm mid},i}$:
    \begin{equation*}
        \loss_{ADE} = \frac{\sum_{i=1}^{t_{f}-1}\|\mathcal{T}_{{\rm mid},i} - \centers_i \|}{t_{f}-1}
    \end{equation*}
    \item Final displacement error between the last predicted center and the final midpoint of the ground truth trajectory:
    \begin{equation*}
        \loss_{FDE} = \| \mathcal{T}_{{\rm mid}, t_{f}-1} - \centers_{t_{f}-1} \|
    \end{equation*}
    \item The midpoint between the current center and previous center $\centers^p_{{\rm mid},i}$ is contained in the current zonotope:
    \begin{equation*}
        \loss_{\rm prev} = \sum_{i=0}^{t_{f}-1} {\rm ReLU}(A_i \cdot \centers^p_{{\rm mid},i} - b_i)
    \end{equation*}
    \item The midpoint between the current center and the next center $\centers^n_{{\rm mid},i}$ is contained in the current zonotope:
    \begin{equation*}
        \loss_{\rm nxt} =  \sum_{i=0}^{t_{f}-1} {\rm ReLU}(A_i \cdot \centers^n_{{\rm mid},i} - b_i)
    \end{equation*}
    \item Regulating the size of the zonotope, by penalizing the norm of the generators such the neural network does not produce excessively large zonotopes that contain the ground truth trajectory:
    \begin{equation*}
        \loss_{G} =  \|  l_{G}[1] - d_1 \| + \| l_{G}[1:]- d_2 \|
    \end{equation*} 
\end{enumerate}
where $d_1$ and $d_2$ are the desired lengths for the generators.
We sum the zonotope shaping losses listed above in a single term $\loss_{\zonotope }$.
Similar to PECNet ~\cite{mangalam2020not}, we use Kullback–Leibler divergence to train the output of the latent encoder, aiming to regulate the divergence between the encoded distribution $\mathcal{N} (\boldsymbol{\mu},\boldsymbol{\sigma})$and the standard normal distribution $\mathcal{N} (0,\boldsymbol{I})$:
\begin{equation*}
    \loss_{KL} = D_{KL}\big(\mathcal{N} (\boldsymbol{\mu},\boldsymbol{\sigma}) \| \mathcal{N} (0,\boldsymbol{I})\big)
\end{equation*}
The network is trained end to end using the following loss function: $\loss = \loss_{KL} + \loss_{\zonotope }$. 

\vspace{0.05in}
\section{Social MPC} \label{sec:SDMPC}
To enable safe navigation in the human-crowded environment, we propose to solve the following optimization problem:
\begin{subequations}
\label{general_mpc}
\begin{align}
\min_{X, U} \quad &\sum_{q = 0}^{N-1}  J(\state, \ctrl) \label{general_mpc_cost} \\
\textrm{s.t.} \quad & \state \nextq = \Phi(\state \currq, \ctrl \currq) \label{general_mpc_dynamics}\\
    & \state_0 = \state_{\rm init}, \;  (\state \currq, \ctrl \currq) \in \set{XU} \currq    \\
  & \state \nextq \in \set{Z} \ego \nextq(\Delta\position\ego_{q}, E \currq) \label{general_mpc_stay_within}\\
  & \set{Z} \ego \nextq(\Delta\position\ego_{q} , E \currq)  \bigcap \set{Z}^{{p_{k \currq}}} \nextq = \emptyset, \; \forall \; k \currq \label{general_mpc_avoid}
\end{align}
\end{subequations}
where the cost~\eqref{general_mpc_cost} is designed to reach the goal and promote social acceptability, subject to the ROM dynamics~\eqref{general_mpc_dynamics} (Sec.~\ref{subsec:ROM}).
Constraint~\eqref{general_mpc_stay_within} requires the ego-agent at the next $(q+1)^{\rm th}$ walking step to stay within the reachable set, while constraint~\eqref{general_mpc_avoid} requires the ego-agent to avoid collision with the pedestrians.
Next, we introduce the kinematics, reachability, and navigation constraints (Sec.~\ref{subsec:kin_const}-\ref{subsec:reachability_and_nav_const}), and finally reformulate the MPC in~\eqref{general_mpc} with a detailed version for implementation (Sec.~\ref{subsec:MPC}).

\subsection{Kinematics Constraints}
\label{subsec:kin_const}
To prevent the LIP dynamics from taking a step that is kinematically infeasible by the Digit robot the following constraint is implemented
\begin{equation}
    \set{XU} \currq = \{(\state \currq, \ctrl \currq) \; | \; \state_{\rm lb} \leq \state \currq \leq \state_{\rm ub} \; \text{and} \; \ctrl_{\rm lb} \leq  \ctrl \currq \leq \ctrl_{\rm ub} \}
    \label{eq:const}
\end{equation}
where $\state_{\rm lb}$ and $\state_{\rm ub}$ are the lower and upper bounds of $\state \currq$ respectively, and $\ctrl_{\rm lb}$ and $\ctrl_{\rm ub}$ are the bounds for $\ctrl \currq$ (See yellow shaded region in Fig.~\ref{fig:LIP}).
The detailed parameters in our implementation are specified in Table~\ref{tab:params}.

\subsection{Reachability and Navigation Safety Constraints}
\label{subsec:reachability_and_nav_const}
To enforce navigation safety (i.e., collision avoidance), we require that Digit remains in the social zonotope $\set{Z} \ego$ and outside of the surrounding pedestrians reachable set $ \hat{\set{Z}}^{{p_{k}}}$. 
\subsubsection{Reachability constrains}

For the robot's CoM to remain inside the desired zonotope for the next walking step $\set{Z} \ego \nextq$, we represent the zonotope using half-space representation as shown in Prop.~\ref{prop:half_space}.
The constraint is reformulated as such:
\begin{equation}
  \max(A \ego \position \ego - b \ego) \leq 0
\end{equation}

\subsubsection{Navigation safety constraint}
For pedestrian collision avoidance, we require that the reachable set of the ego-agent does not intersect with that of the pedestrians for the corresponding step.
Therefore, we create a new zonotope for the ego-agent as Minkowski sum of the ego-agent's zonotope and the pedestrian's zonotope centered around the ego-agent $\set{Z} \mink =\mathscr{Z}(\boldsymbol{c} \ego, [{G}^{{\rm ego}} \; G \ped])$ to check for collision with the pedesrians' zonotope following Prop.~\ref{prop:intersection}.
We then represent $\set{Z} \mink$ using half-space representation and require that the pedestrian is outside the combined set:
\begin{equation}
    \max(A \mink \position_k - b \mink) > 0
\end{equation}

\subsection{Cost Function}
\label{subsec:cost}

The MPC cost function is designed to drive the ROM state to a goal location $\set{G}$.
The terminal cost penalizes the distance between the current ROM state and the global goal state $\set{G}$.
\begin{equation}
    J_N(\state_{N}) = \| \state_{N} - \state_{\set{G}} \|^2_{W_1} + \| \theta_{N} - \theta_{\set{G}} \|^2_{W_2}
\end{equation}
where $\state_{\set{G}} = (\set{G}, v_{\rm terminal})$, and $\theta_\set{G}$ is the angle between the ego-agent's current position and the final goal location.

\subsection{MPC Reformulation with Ego-agent Social Network}
\label{subsec:MPC}

According to the aforementioned costs and constraints for implementation, we reformulate our Ego-agent Social Network MPC (ESN-MPC) shown in~\eqref{general_mpc} as follows:
\begin{subequations}
\label{eq:problem}
\begin{align}
\min_{X, U} \quad &\sum_{\mathfrak{q}=0}^{N-1}   J_N(\state_{N})  \\
\textrm{s.t.} \quad & \state \nextq = \Phi(\state \currq, \ctrl \currq)\\
    & \state_0 = \state_{\rm init}, \;  (\state \currq, \ctrl \currq) \in \set{XU} \currq    \\
  &  \max(A \ego \nextq \position \ego \nextq - b \ego \nextq) \leq 0\\
  &  \max(A \mink \nextq \position_{k \nextq} - b \mink \nextq) > 0, \; \forall \; k \currq  
\end{align}
\end{subequations}

\vspace{0.05in}
\section{Implementation and Results}
\label{sec:implementation}

\subsection{Training}
The social path planner module introduced in Sec.~\ref{sec:social_zono_net} was trained on the UCY~\cite{lerner2007crowds} and ETH~\cite{pellegrini2009walk} crowd datasets with the common leave-one-out approach, reminiscent of prior studies~\cite{salzmann2020trajectron++,mangalam2020not, gupta2018social}.

The models were trained on a data set that excludes UNIV from the training examples.
We employ a historical trajectory observation $\mathcal{T}^{p_k}_{[-8, 0]}$ and a prediction horizon $\hat{\mathcal{T}}^{\rm ego}_{[0, 8]}$, each spanning a duration of $8$ timesteps ($3.2$ s) and only consider neighboring pedestrians that are within a radius of $4$ m. The network architecture details are shown in Table~\ref{tab:network_arch}.

\begin{table}
    \centering
    \caption{Network architecture parameters}
    \begin{tabular}{|c|c|}
       \hline
        \multicolumn{2}{|c|}{Pedestrian Prediction Network} \\
        \hline
       $E_{\rm ped}$ & $16\rightarrow32\rightarrow16$  \\
       $E_{\rm end}$ & $2 \rightarrow 8 \rightarrow 16$  \\
       $E_{\rm nxt}$ & $2 \rightarrow 32 \rightarrow 16$  \\
       $P_{\rm future}$  & $50 \rightarrow 32 \rightarrow 16 \rightarrow 32 \rightarrow 70$ \\
        $E_{\rm latent}$ & $48 \rightarrow 8 \rightarrow 16 \rightarrow 32$  \\
        $D_{\rm latent}$ & $48 \rightarrow 32 \rightarrow 16 \rightarrow 32 \rightarrow 2$ \\
         \hline
         \hline
        \multicolumn{2}{|c|}{Ego-agent Social Network} \\
        \hline
       $E_{\rm goal}$ & $2\rightarrow 8 \rightarrow 16 \rightarrow2$  \\
       $E_{\rm future}$ & $16 \rightarrow 64 \rightarrow 32 \rightarrow 16$  \\
       $E_{\rm nxt}$ & $2 \rightarrow 64 \rightarrow 32 \rightarrow 2$  \\
       $E_{\rm traj}$  & $16 \rightarrow 64 \rightarrow 32 \rightarrow 16$ \\
        $E_{\rm latent}$ & $36 \rightarrow 8 \rightarrow 50 \rightarrow 16$  \\
        $D_{\rm latent}$ & $36 \rightarrow 128 \rightarrow 64 \rightarrow 128 \rightarrow 70$ \\
         \hline
    \end{tabular}

    \label{tab:network_arch}
\end{table}

\subsection{Pedestrian Simulation}
We use SGAN (Social Generative Adversarial Network), a state-of-the-art human trajectory model,  for simulating pedestrians \cite{gupta2018social}.
SGAN is specifically designed to grasp social interactions and dependencies among pedestrians.
It considers social context, including how people influence each other and move in groups.
This is important for creating realistic simulations of pedestrian motion.
Employing a different prediction model ensures a fair evaluation by eliminating any inherent advantages of our proposed method\cite{schaefer2021leveraging}. 

In our simulation framework, SGAN incorporates both the historical trajectories of pedestrians and the trajectory of the ego agent.
This approach enhances the realism of the simulation by accounting for the interaction between the ego-agent and pedestrians within the environment.

\subsection{Testing Environment Setup}
\label{subsubsec:test_env}
The environment for all the following tests is an open space of $14 \times 14$ m$^2$ as shown in Fig.~\ref{fig:high-level} and Fig.~\ref{fig:frames}, with randomly generated pedestrians' initial trajectory. We test with $5$, $15$, and $30$ pedestrians in the environment. 
The goal location is $\mathcal{G}=(10,10)$ m, and the ego-agent starting position is uniformly sampled along the $y$-axis as such $\boldsymbol{x}_0 = (0, \mathcal{U}_{[0, 13]},0)$ with $\theta_0 = 0$.
The MPC is solved with a planning horizon of $N=4$, and ESN-MPC parameters are included in Table.~\ref{tab:params}.
Simulations and training are done using a 16-core Intel Xeon W-2245 CPU and an RTX-5000 GPU with 64 GB of memory.

\begin{table}
    \centering
    \caption{ESN-MPC Parameters}
    \begin{tabular}{|c|c|c|c|}
    \hline
        parameter & value & parameter & value \\
    \hline
       $u^{\Delta \theta}_{\rm ub}$ & $15^\circ$ & $u^{\Delta \theta}_{\rm lb}$ & $-15^\circ$\\
    \hline
        $u^{f}_{\rm ub}$ & $0.4$ m & $u^{f}_{\rm lb}$ & $-0.1$ m\\
    \hline
       $d_1$  & $0.1$ & $d_2$ & $0.005$ \\
    \hline
        $v_{\rm terminal}$ & $0$ m/s & $n_{G}$ & $4$ \\
    \hline
        $W_1$ & $3$ & $W_2$ & $1$\\
    \hline
    \end{tabular}
    
    \label{tab:params}
    
\end{table}

\subsection{Low-level Full-Body Control}
At the low level we use the Angular momentum LIP planner introduced in~\cite{Gong2022AngularMomentum}, and a Digit's passivity controller~\cite{sadeghian2017passivity} with ankle actuation which we have previously shown to exhibit desirable ROM tracking results~\cite{shamsah2023integrated}.
Here we set the desired walking step time and the desired lateral step width to be fixed at $0.4$ s and $0.4$ m, respectively.
\subsection{Results and Discussion}\label{sec:results}

\begin{figure}[t]
\centerline{\includegraphics[width=.8\columnwidth]{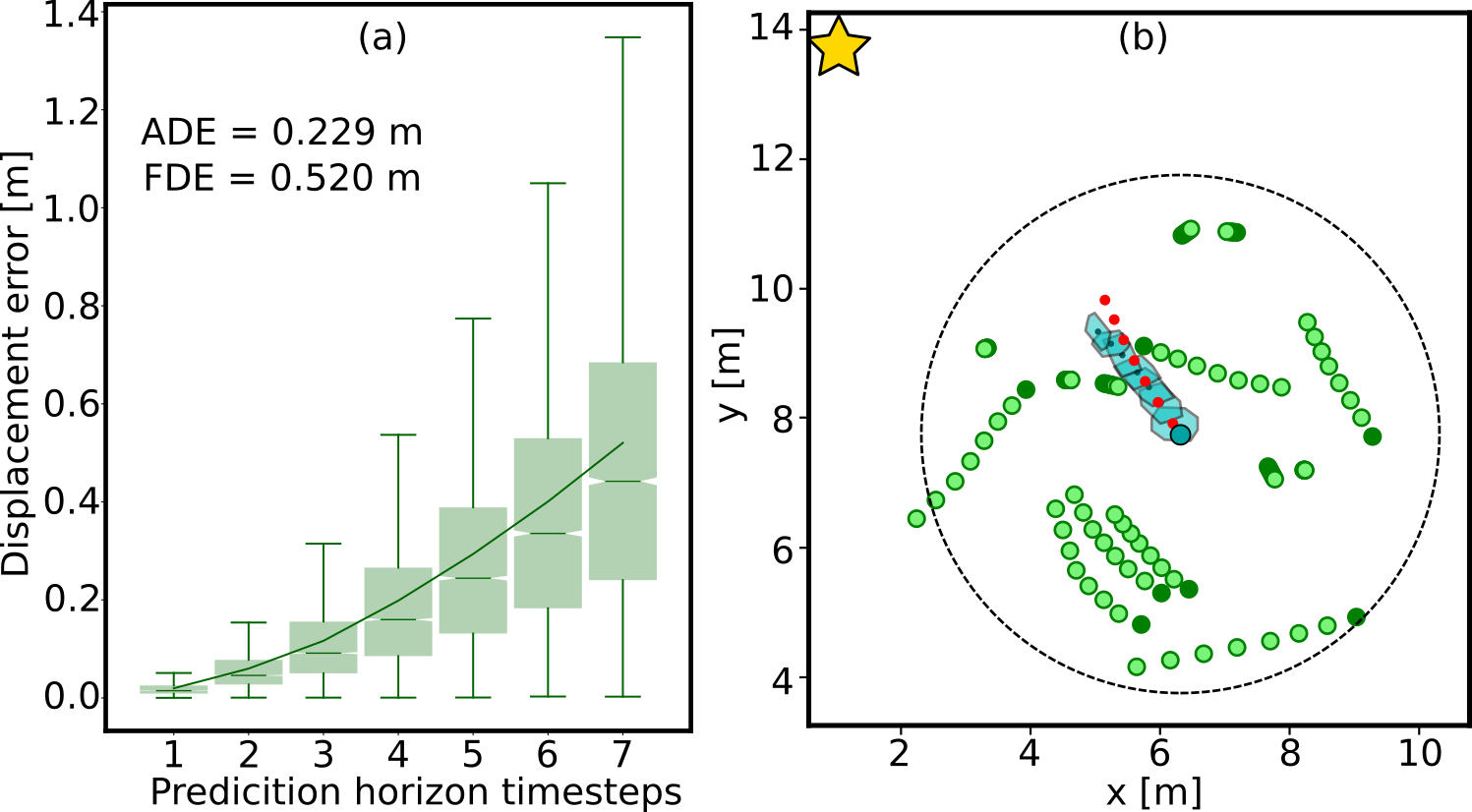}}
\caption{Quantitative (a) and qualitative (b) results of ESN. (a) Shows the displacement error between the prediction of ESN $\boldsymbol{c}^{\rm ego}$ (centers of cyan zonotopes in (b)) and the ground truth data $\mathcal{T}^{\rm ego}_{\rm mid} $ (red dots in (b)). (b) shows a snapshot of ESN output, where the ego-agent's predicted zonotopes (cyan) contain the ground truth ego-agent data (red). ESN is conditioned on the goal position (yellow $\star$) and surrounding pedestrian future trajectories (green). The data is collected based on the UNIV dataset with $7831$ unique frames. The solid line in (a) shows the average displacement error at each prediction horizon.}
\label{fig:ADE}
\end{figure}

\begin{figure}[t]
\centerline{\includegraphics[width=.8\columnwidth]{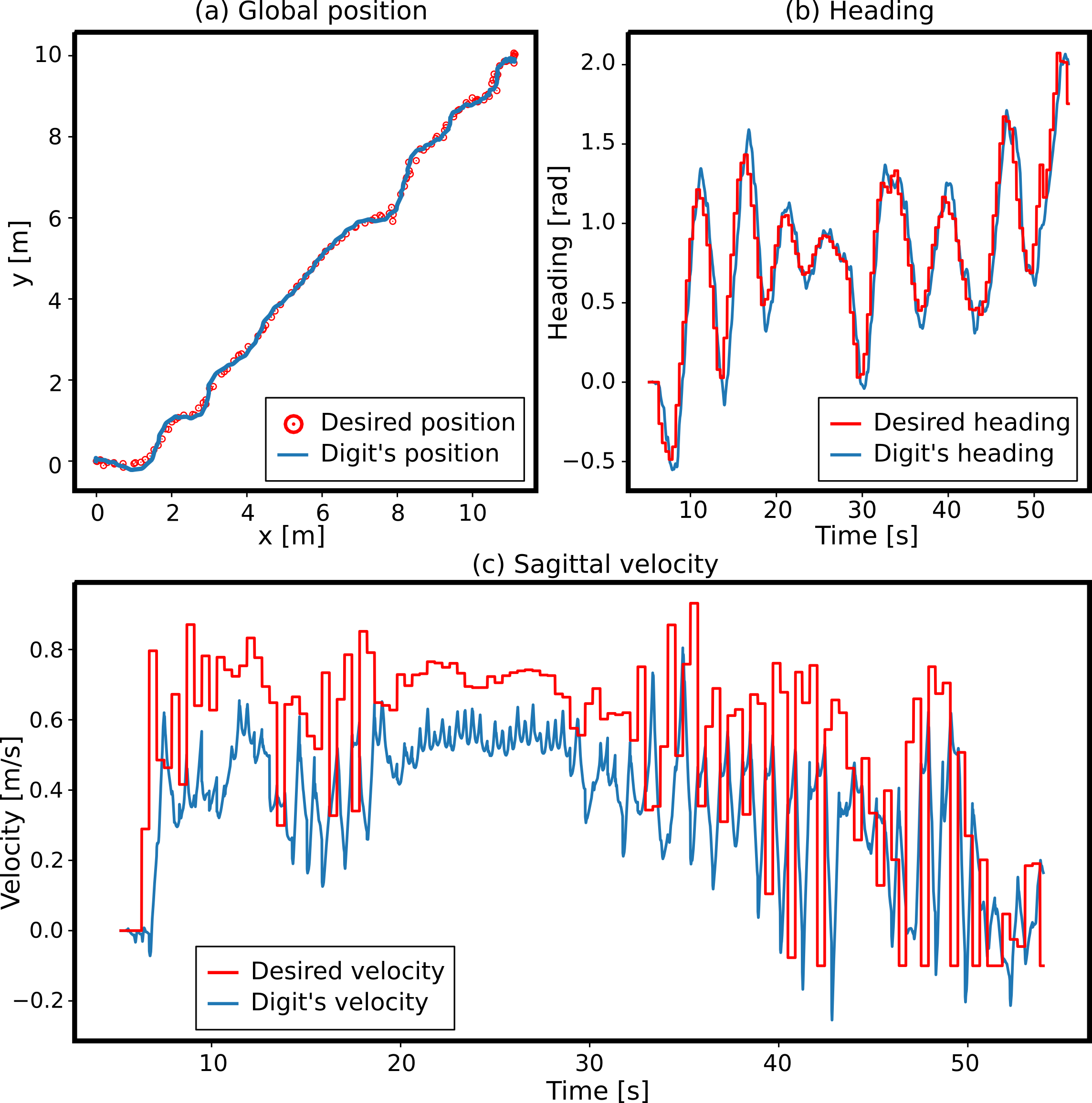}}
\caption{Full-order simulation results of Digit tracking the desired trajectory from ESN-MPC. (a) shows Euclidean position tracking, (b) shows heading tracking, and (c) shows the sagittal velocity tracking in local coordinates.}
\label{fig:full_body}
\end{figure}

\begin{figure*}[t]
\centerline{\includegraphics[width=.8\textwidth]{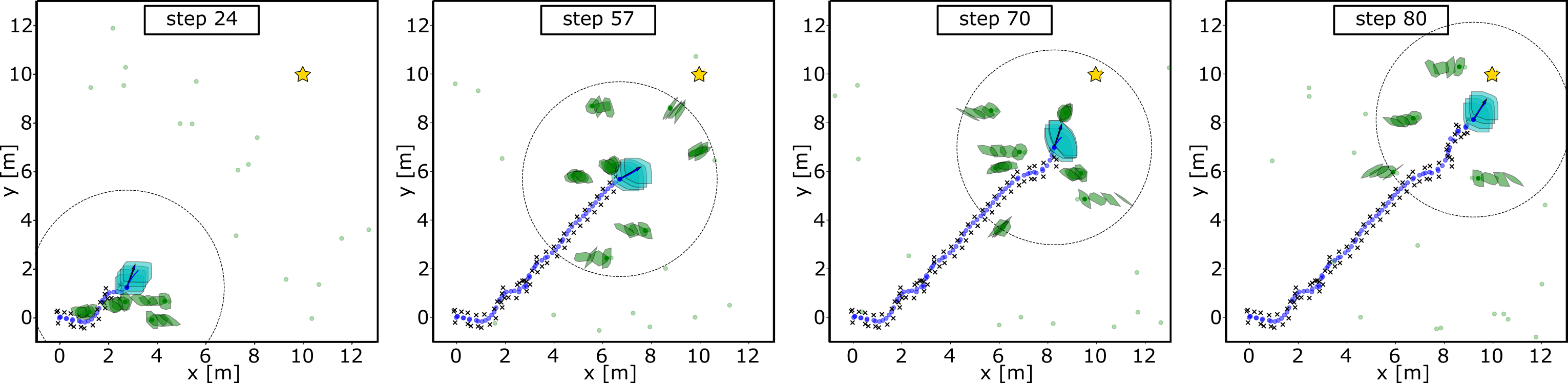}}
\caption{Storyboard snapshots of ESN-MPC trajectory at different walking steps. The ego-agent (cyan) successfully reaches the goal (yellow $\star$) while avoiding pedestrians (green). The dashed circle is the sensory radius of the ego-agent. Cayan dots represent the CoM of ROM, and black $\times$ is the desired foot placement.
Green dots are unobserved pedestrians.}
\label{fig:frames}
\end{figure*}
\begin{figure*}[t]
\centerline{\includegraphics[width=.98\textwidth]{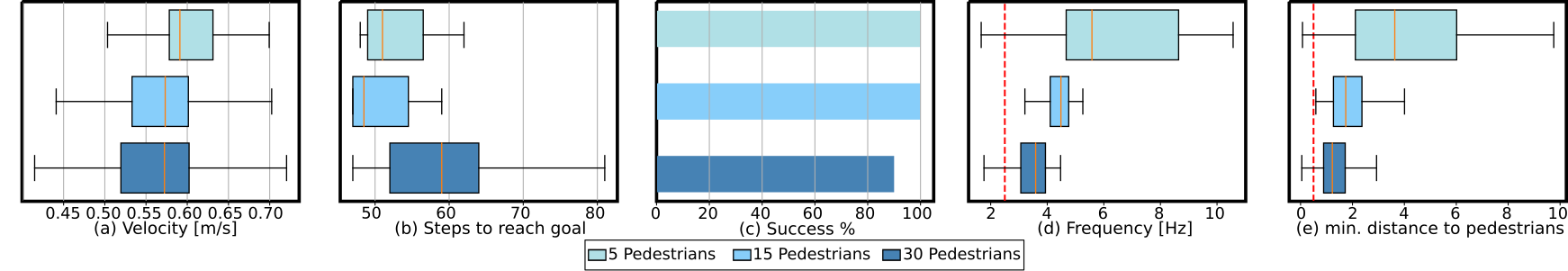}}
\caption{ESN-MPC results: (a) velocity, (b) number of walking steps to reach within $1$ m of the goal, (c) success rate, (d) frequency, and (e) minimum distance to pedestrians. The data consists of 10 different trials with random initial conditions and a fixed goal location (see Sec.~\ref{subsubsec:test_env}). Each trail is $100$ walking steps. The velocity data is collected before reaching the goal, to avoid collecting a stop velocity.  
Success represents reaching within $1$ m of the goal in $100$ steps.  The dashed red line in (d) is the minimum required frequency for full-order Digit simulation. The dashed red line (e) represents $d$ in navigation safety (Definition.~\ref{def:nav_safety}).}
\label{fig:results}
\vspace{-0.15in}
\end{figure*}

In Fig.~\ref{fig:ADE}(a) we show that ESN produces an ADE$=0.229$ m over the prediction horizon of $7$ timesteps\footnote{The prediction horizon timesteps is $7$ and not $8$, since the displacement error is calculated based on the middle points $\mathcal{T}^{\rm ego}_{\rm mid}$ of $\mathcal{T}^{\rm ego}_{[0, 8]}$}, and a Final Displacement Error (FDE)$=0.52$ m. Fig.~\ref{fig:ADE}(b) shows a snapshot of the ESN social zonotope output $\set{Z} \ego$ (cyan) compared to the ground truth data $\mathcal{T}^{\rm ego}_{\rm mid}$ shown in red.

Fig.~\ref{fig:full_body} shows the tracking performance of integrating ESN-MPC with the low-level full-body controller~\cite{Gong2022AngularMomentum,sadeghian2017passivity,shamsah2023integrated}.
We show the global Euclidean position tracking Fig.~\ref{fig:full_body}(a), heading angle tracking in Fig.~\ref{fig:full_body}(b), and local sagittal velocity tracking in~Fig.~\ref{fig:full_body}(c). Fig.~\ref{fig:high-level} and Fig.~\ref{fig:frames} show snapshots of the resultant trajectory at different walking steps. A video of the simulations is \href{https://rb.gy/z9z3zj}{\nolinkurl{https://rb.gy/z9z3zj}}. 

In Fig.~\ref{fig:results}(a), all three crowd densities produce relatively similar median velocities. At lower crowd density the velocity is more consistent.
As expected, Fig.~\ref{fig:results}(b) shows that in less crowded areas, the ego-agent can reach the goal in fewer steps. With $30$ pedestrians in the environment, it took more steps on average to reach the goal while maintaining a relatively similar velocity to the environments with fewer crowds (see Fig.~\ref{fig:results}(a)). This indicates that our framework can predict the future trajectory of the surrounding pedestrians, and is not required to come to a sudden stop.
ESN-MPC produces a consistent and predictable behavior for the ego-agent.
Predictability of the ego-agent behavior in a social context is desirable by pedestrians as it is perceived to be less disruptive.
With $5$ and $15$ pedestrians in the environment our framework produced a $100 \%$ success rate by reaching the goal in $100$ walking steps, while it managed a $90 \%$ success rate with $30$ pedestrians as shown in Fig.~\ref{fig:results}(c).
Due to the larger number of constraints, the time it takes to solve ESN-MPC decreases with increasing the number of pedestrians (See Fig.~\ref{fig:results}(d)).
However, even with $30$ pedestrians, the median of the frequency is higher than the required minimum for Digit implementation as indicated by the dashed red line in Fig.~\ref{fig:results}(d).
Finally, ESN-MPC can maintain a safe distance to the pedestrians in all three testing environments as indicated in Fig.~\ref{fig:results}(e).

\vspace{0.05in}
\section{Conclusion}\label{sec:conclusion}
This study introduced a novel framework for bipedal robot navigation in human environments, addressing a significant gap in the field of locomotion navigation.
The proposed framework, which comprises the Pedestrian Prediction Network (PPN) and the Ego-agent Social Network (ESN), leverages zonotopes for efficient reachability-based planning and collision checking. Integrating ESN with MPC for step planning for Digit showed promising results for safe navigation in social environments. Future work will explore exploiting the zonotope parameterization for modeling ROM and full-order model discrepancies, introducing locomotion-specific losses to ESN training to facilitate safe hardware experimentation, and quantifying social acceptability.

\appendices



\bibliographystyle{IEEEtran}
\bibliography{lidar.bib}

\end{document}